# Avian Influenza (H5N1) Warning System using Dempster-Shafer Theory and Web Mapping

Andino Maseleno, Md. Mahmud Hasan
Department of Computer Science, Faculty of Science, Universiti Brunei Darussalam
Jalan Tungku Link, Gadong BE 1410, Negara Brunei Darussalam
E-mail: andinomaseleno@yahoo.com, mahmud.hasan@ubd.edu.bn

*Abstract*— **Based on Cumulative Number of Confirmed Human Cases of Avian Influenza (H5N1) Reported to World Health Organization (WHO) in the 2011 from 15 countries, Indonesia has the largest number death because Avian Influenza which 146 deaths. In this research, the researcher built a Web Mapping and Dempster-Shafer theory as early warning system of avian influenza. Early warning is the provision of timely and effective information, through identified institutions, that allows individuals exposed to a hazard to take action to avoid or reduce their risk and prepare for effective response. In this paper as example we use five symptoms as major symptoms which include depression, combs, wattle, bluish face region, swollen face region, narrowness of eyes, and balance disorders. Research location is in the Lampung Province, South Sumatera. The researcher reason to choose Lampung Province in South Sumatera on the basis that has a high poultry population. Geographically, Lampung province is located at $103^040'$ to $105^050'$ East Longitude and $6^045'$ – $3^045'$ South latitude, confined with: South Sumatera and Bengkulu on North Side, Sunda Strait on the Side, Java Sea on the East Side, Indonesia Ocean on the West Side. Our approach uses Dempster Shafer theory to combine beliefs in certain hypotheses under conditions of uncertainty and ignorance, and allows quantitative measurement of the belief and plausibility in our identification result. Web Mapping is also used for displaying maps on a screen to visualize the result of the identification process. The result reveal that avian influenza warning system has successfully identified the existence of avian influenza and the maps can be displayed as the visualization.**

## I. INTRODUCTION

Historically, outbreaks of avian influenza first occurred in Italy in 1878, when it was a lot of dead birds [1]. Then came another outbreak of Avian Influenza in Scotland in 1959 [2]. The virus that causes Avian Influenza in Italy and Scotland are the current strain of H5N1 virus appeared again attacked poultry and humans in various countries in Asia, including Indonesia, which caused many deaths in humans [3].

Avian influenza virus H5N1, which has been limited to poultry, now has spread to migrating birds and has emerged in mammals and among the human population. It presents a distinct threat of a pandemic for which the World Health Organization and other organizations are making preparations. In 2005, the World Health Assembly urged its Member States to develop national preparedness plans for pandemic influenza [4]. Developing countries face particular planning and other challenges with pandemic preparedness as there may be a higher death rate in developing countries compared with more developed countries [5]. In this research, we use chicken as research object because chicken population has grown very fast in Lampung Province at 2009, native chicken population around 11,234,890, broiler population around 15,879,617, layer population around 3,327,847. Lampung Province has been divided into 10 regencies, 204 districts and 2279 villages with area of 3,528,835 hectare [6].

To overcome avian influenza required as an early warning system of avian influenza. International Strategy for Disaster Reduction (ISDR) defines early warning as the provision of timely and effective information, through identified institutions, that allows individuals exposed to a hazard to take action to avoid or reduce their risk and prepare for effective response [7]. The problems that exists are how an early warning system to the identification of avian influenza and also the location where the diseases can give benefit for poultry husbandry and animal health agencies as decision makers in animal husbandry and animal health problems.

The remainder is organized as follows. The Web mapping is briefly reviewed in Section 2. Section 3 details the proposed Dempster-Shafer Theory. Architecture of Avian Influenza Warning System is detailed in Section 4. The experimental results are presented in Section 5, and final remarks are concluded in Section 6.

## II. WEB MAPPING

Web mapping is the process of designing, implementing, generating and delivering maps on the World Wide Web and its product. While web mapping primarily deals with technological issues, web cartography additionally studies theoretic aspects: the use of web maps, the evaluation and optimization of techniques and workflows, the usability of web maps, social aspects, and more. Web Geographic Information System (GIS) is similar to web mapping but with an emphasis on analysis, processing of project specific geo data and exploratory aspects [8].

The web mapping server is the engine behind the maps [9]. The mapping server or web mapping program needs to be configured to communicate between the web server and assemble data layers into an appropriate image. A map is not possible without some sort of mapping information for display. Mapping data is often referred to as spatial or geospatial data and can be used in an array of desktop mapping programs or web mapping servers. Mapping data in the avian influenza warning system uses spatial and non-spatial data in ArcView format. Table 1 shows the mapping data.

TABLE I. DATA MAPPING

| Data Type | Data Name | Description |
|---|---|---|
| Spatial | Province Regencies District Desa | Spatial Data digitized on screen with ArcView |
| Non-Spatial | Province Table Regency Table District Table Village Table | Data tabulated into flat table of which follow data spatial |

Information displayed spatial data is the appearance of a map based on the selected layer, while the non-spatial data is information support from spatial data being displayed.

## III. DEMPSTER-SHAFER THEORY

The Dempster-Shafer theory was first introduced by Dempster [10] and then extended by shafer [11], but the kind of reasoning the theory uses can be found as far back as the seventeenth century. This theory is actually an extension to classic probabilistic uncertainty modeling. Whereas the Bayesian theory requires probabilities for each question of interest, belief functions allow us to base degrees of belief for on question on probabilities for a related question. In terms of previous work using Dempster-Shafer theory to estimate stand regeneration maps [12], [13]. Actually, according to researchers knowledge, Dempster-Shafer theory of evidence has never been used for built an early warning system of avian influenza merging with Web Mapping. The advantages of the Dempster-Shafer theory as follows:
1. It has the ability to model information in a flexible way without requiring a probability to be assigned to each element in a set,
2. It provides a convenient and simple mechanism (Dempster's combination rule) for combining two or more pieces of evidence under certain conditions.
3. It can model ignorance explicitly.
4. Rejection of the law of additivity for belief in disjoint propositions.

Avian influenza warning system using Dempster-Shafer theory in the decision support process. Flowchart of avian influenza warning system shown in Figure 1.

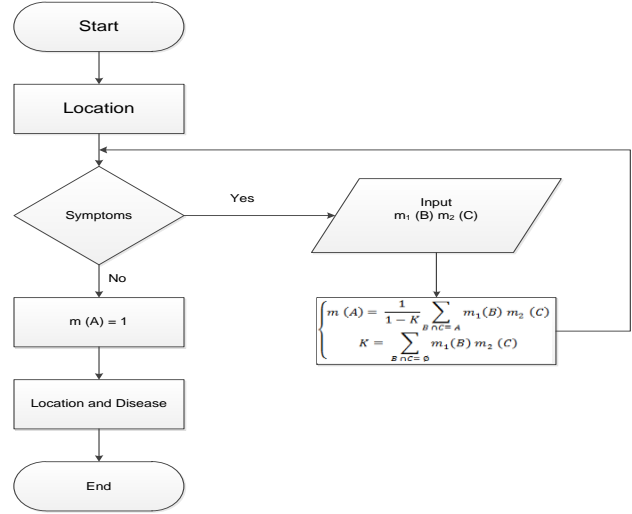

Figure 1. Flowchart of Avian Influenza Warning System

The consultation process begins with selecting the location and symptoms. If there are symptoms then will calculate, The Dempster Shafer theory provides a rule to combine evidences from independent observers and into a single and more informative hint. Evidence theory is based on belief function and plausible reasoning. First of all, we must define a frame of discernment, indicated by the sign $\Theta$. The sign $2^\Theta$ indicates the set composed of all the subset generated by the frame of discernment. For a hypothesis set, denoted by A, $m(A) \rightarrow [0,1]$.

$m(\emptyset) = 0$

$$\sum_{A \in 2^\Theta} m(A) = 1 \quad (1)$$

$\emptyset$ is the sign of an empty set. The function m is the basic probability assignment. Dempster's rule of combination combines two independent sets of mass assignments.

$$(m_1 \oplus m_2)(\emptyset) = 0 \quad (2)$$

$$(m_1 \oplus m_2)(A) = \frac{1}{1-K} \sum_{B \cap C = A} m_1(B) m_2(C) \quad (3)$$

$$K = \sum_{B \cap C = \emptyset} m_1(B) m_2(C) \quad (4)$$

$m(A), m_1(B), m_2(C) \rightarrow [0,1], A \neq \emptyset$

The result is that avian influenza warning system successfully identifying disease and displaying location to visualize the result of identification process.
Symptoms:

1. Depression
2. Combs, wattle, bluish face region
3. Swollen face region
4. Narrowness of eyes
5. Balance disorder

*A. Symptom 1*

Depression is a symptom of Avian Influenza (AI), Newcastle Disease (ND), Fowl Cholera (FC), Infectious

Bronchitis respiratory form (IBRespi), Infectious Bronchitis reproduction form (IBRepro), and Swollen Head Syndrome (SHS). The measures of uncertainty, taken collectively are known in Dempster Shafer Theory terminology as a ``basic probability assignment'' (bpa). Hence we have a bpa, say $m_1$ of 0.7 given to the focal element {AI, ND, FC, IBRespi, IBRepro, SHS} in example, $m_1$({AI, ND, FC, IBRespi, IBRepro, SHS}) = 0.7, since we know nothing about the remaining probability it is allocated to the whole of the frame of the discernment in example, $m_1$({AI, ND, FC, IBRespi, IBRepro, SHS}) = 0.3, so:
$m_1${AI, ND, FC, IBRespi, IBRepro, SHS} = 0.7
$m_1${Θ} = 1 - 0.7 = 0.3

B. *Symptom 2*

Combs, wattle, bluish face region are symptoms of Avian Influenza with a bpa of 0.9, so that:
$m_2${AI} = 0.9
$m_2$ {Θ} = 1 − 0.9 = 0.1

With the symptoms comb, wattle, bluish face region then required to calculate the new bpa values for some combinations ($m_3$). Combination rules for the $m_3$ can be seen in the table 1.

TABLE II. COMBINATION OF SYMPTOM 1 AND SYMPTOM 2

|  |  | {AI} | 0.9 | Θ | 0.1 |
|---|---|---|---|---|---|
| {AI, ND, FC, IBRespi, IBRepro, SHS} | 0.7 | {AI} | 0.63 | {AI, ND, FC, IBRespi, IBRepro, SHS} | 0.07 |
| Θ | 0.3 | {AI} | 0.27 | Θ | 0.03 |

$$m_3 (AI) = \frac{0.63 + 0.27}{1 - 0} = 0.9$$

$$m_3 (AI, ND, FC, IBRespi, IBRepro, SHS) = \frac{0.07}{1-0} = 0.07$$

$$m_3 (\Theta) = \frac{0.03}{1-0} = 0.03$$

C. *Symptom 3*

Swollen face region is a symptom of Avian Influenza, Newcastle Disease, Fowl Cholera with a bpa of 0.83, so that
$m_4$ {AI, ND, FC} = 0.83
$m_4$ (Θ) = 1 − 0.83 = 0.17

With the symptom swollen face region then required to calculate the new bpa values for each subset.

TABLE III. COMBINATION OF SYMPTOM 1, SYMPTOM 2, AND SYMPTOM

|  |  | {AI, ND, FC} | 0.83 | Θ | 0.17 |
|---|---|---|---|---|---|
| {AI} | 0.9 | {AI} | 0.747 | {AI} | 0.153 |
| {AI, ND, FC, IBRespi, IBRepro, SHS} | 0.07 | {AI, ND, FC} | 0.0581 | {AI, ND, FC, IBRespi, IBRepro, SHS} | 0.0119 |
| Θ | 0.03 | {AI, ND, FC} | 0.0249 | Θ | 0.0051 |

D. *Symptom 4*

Narrowness of eyes is a symptom of Swollen Head Syndrome with a bpa of 0.9, so that:
$m_6$ (SHS) = 0.9
$m_6$ (Θ) = 1 − 0.9 = 0.1

With the symptom narrowness of eyes then required to calculate the new bpa values for each subset with bpa $m_7$. Combination rules for $m_7$ can be seen in Table 4.

TABLE IV. COMBINATION OF SYMPTOM 1, SYMPTOM 2, SYMPTOM 3, AND SYMPTOM 4

|  |  | {SHS} | 0.9 | Θ | 0.1 |
|---|---|---|---|---|---|
| {AI} | 0.9 | Θ | 0.81 | {AI} | 0.09 |
| {AI, ND, FC} | 0.083 | Θ | 0.0747 | {AI, ND, FC} | 0.0083 |
| {AI, ND, FC, IBRespi, IBRepro, SHS} | 0.0119 | {SHS} | 0.01071 | {AI, ND, FC, IBRespi, IBRepro, SHS} | 0.00119 |
| Θ | 0.0051 | {SHS} | 0.00459 | Θ | 0.00051 |

$$m_7 (SHS) = \frac{0.01071 + 0.00459}{1 - (0.81 + 0.0747)} = 0.13270$$

$$m_7 (AI) = \frac{0.09}{1 - (0.81 + 0.0747)} = 0.78057$$

$$m_7 (AI, ND, FC) = \frac{0.0083}{1 - (0.81 + 0.0747)} = 0.07199$$

$$m_7 (AI, ND, FC, IBRespi, IBRepro, SHS) = \frac{0.00119}{1 - (0.81 + 0.0747)} = 0.01032$$

$$m_7 (\Theta) = \frac{0.00051}{1 - (0.81 + 0.0747)} = 0.00442$$

E. *Symptom 5*

Balance disorders is a symptom of Newcastle Diseases and Swollen Head Syndrome with a bpa of 0.6, so that:
$m_8$ {ND,SHS} = 0.6
$m_8$ {Θ} = 1 - 0.6 = 0.4

With the symptom balance disorders will be required to calculate the new bpa values for each subset with $m_9$ bpa. Combination rules for the $m_9$ can be seen in Table 5.

TABLE V. COMBINATION OF SYMPTOM 1, SYMPTOM 2, SYMPTOM 3, SYMPTOM 4, AND SYMPTOM 5

|  |  | {ND, SHS} | 0.6 | Θ | 0.4 |
|---|---|---|---|---|---|
| {SHS} | 0.13270 | {SHS} | 0.07962 | {SHS} | 0.05308 |
| {AI} | 0.78057 | Θ | 0.46834 | {AI} | 0.31222 |
| {AI, ND, FC} | 0.07199 | {ND} | 0.04319 | {AI, ND, FC} | 0.02880 |
| {AI, ND, FC, IBRespi, IBRepro, SHS } | 0.01032 | {ND, SHS} | 0.00619 | {AI, ND, FC, IBRespi, IBRepro, SHS } | 0.00413 |
| Θ | 0.00442 | {ND, SHS} | 0.00265 | Θ | 0.00177 |

$$m_9(SHS) = \frac{0.07962 + 0.05308}{1 - 0.46834} = 0.24960$$

$$m_9(AI) = \frac{0.31222}{1 - 0.46834} = 0.58725$$

$$m_9(ND) = \frac{0.04319}{1 - 0.46834} = 0.08124$$

$$m_9(ND, SHS) = \frac{0.00619 + 0.00265}{1 - 0.46834} = 0.01663$$

$$m_9(AI, ND, FC) = \frac{0.02880}{1 - 0.46834} = 0.05417$$

$$m_9(AI, ND, FC, IBRespi, IBRepro, SHS) = \frac{0.00413}{1 - 0.46834} = 0.00777$$

$$m_9(\Theta) = \frac{0.000232}{1 - 0.061038} = 0.00025$$

The most highly bpa value is the $m_9$ (AI) that is equal to 0.58725 which means the possibility of a temporary diseases with symptoms of depression, comb, wattle, bluish face region, swollen region face, narrowness of eyes, and balance disorders is the Avian influenza (H5N1).

## IV. AVIAN INFLUENZA WARNING SYSTEM

Figure 2 shows architecture of Avian Influenza Warning System which is fusion between the Dempster-Shafer theory and Web Map Application.

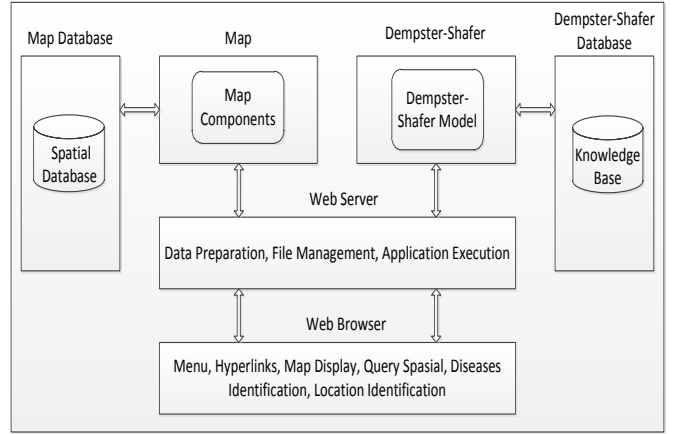

Figure 2. Architecture of Avian Influenza Warning System

The development of Avian Influenza Warning System to begin with:
1. Preparation which includes installing and setting programs.
2. Preparation of attributes data and spatial data that already exist so it can be used by the program application.
3. Database design with MySQL. At this stage the databases design regarding the depiction relationships between entities using diagram of relationships between entities (entity relationship diagrams).
4. Designing Web pages using PHP. This web page then connected to the database and spatial attributes that have been established to form a network of database.
5. Designing a Web-based applications used to access the network database that has been formed. The relationship between applications with network data base that has shaped overall applications to manage information to be displayed.

Applications that have been built then tested by using samples of existing data to determine whether application is already running technically. When test results are less satisfactory then conducted a review the stages of database design, web page design, or web application design that have been made to do repairs required.

## V. IMPLEMENTATION

Avian Influenza warning system applying web mapping technology using MapServer software. Map data obtained from digitized by ArcView for creating and editing dataset. The implementation process is done after design and scope of the system have been analyzed. Writing map file using Macromedia Dreamweaver. Macromedia Dreamweaver for writing map file that later on as the main file from system configuration and layout as well. Spatial and non spatial data designed using ArcView. Each data is designed to accommodate operation at the layer level, either single layer or multi layer.

The following will be shown the working process of expert system in diagnosing a case. The consultation process begins with selecting the location and symptoms found on the list of symptoms. In the cases tested, a known symptoms are depression, comb, wattle, bluish-colored facade region, region of the face swollen, eyes narrowed and balance disorders. The consultation process can be seen in Figure 3 and Figure 4.

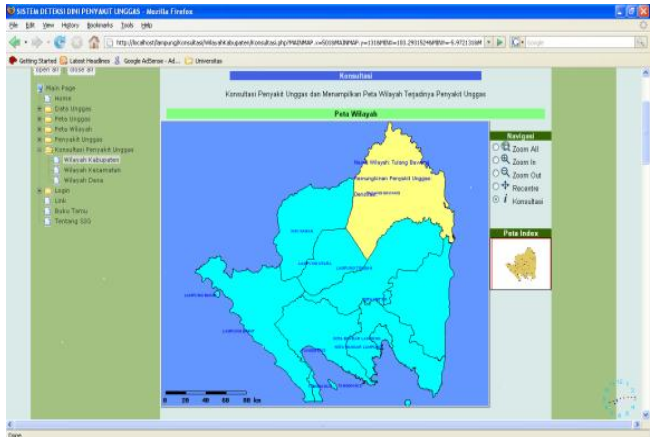
Figure 3. Selecting Region of Consultation

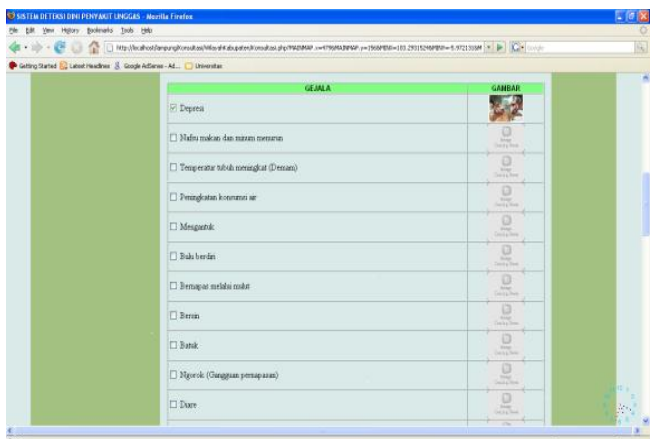
Figure 4. Symptoms Selection

In the case of depression, comb, wattle and region of the face bluish, region of the face swollen, eyes narrowed and lachrymal glands swollen. The result of consultation is avian influenza with bpa value equal to 0.587275693312. Figure 5 and 6 shows the result of consultation and the region map of consultation result.

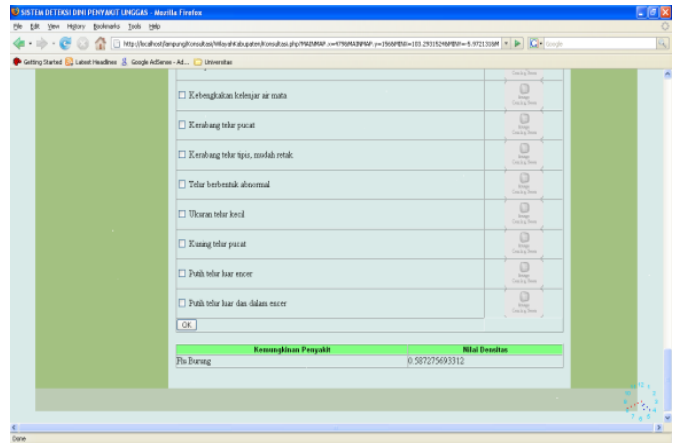
Figure 5. The Result of Consultation

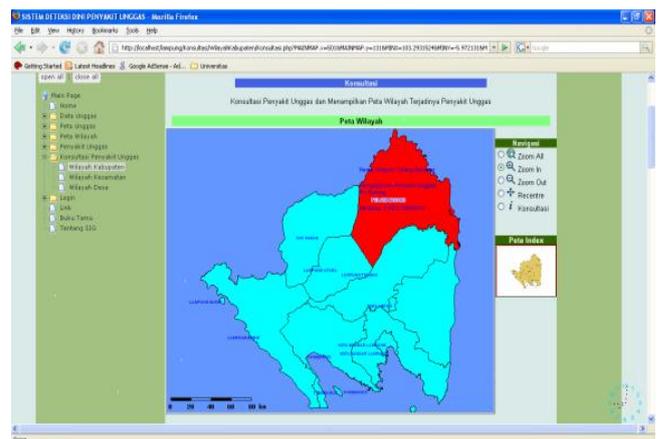
Figure 6. The Region Map of Consultation Result

## VI. CONCLUSION

Identification of avian influenza can be performed using Dempster-Shafer Theory and Web Mapping. In this paper as example we use five symptoms as major symptoms are depression, combs, wattle, bluish face region, swollen face region, narrowness of eyes, and balance disorders. The knowledge is uncertain in the collection of basic events can be directly used to draw conclusions in simple cases, however, in many cases the various events associated with each other. Knowledge based is to draw conclusions, it is derived from uncertain knowledge. Application is built to display the map of the region by the villages, districts, and regencies or municipalities. Map has been used to solve the problem of Avian Influenza. Map can be used to integrate spatial data and descriptive data, early warning system can use map as a tool. Web Mapping and Dempster Shafer theory can be constructed as an early warning system of avian influenza, with avian influenza identification process and display the map of the region identification. This research can be an alternative in addition to direct consultation with doctor and to find out quickly location of avian influenza.